\documentclass[11pt]{article}

\usepackage[preprint]{acl}

\usepackage{times}
\usepackage{latexsym}

\usepackage[T1]{fontenc}
\usepackage[utf8]{inputenc}

\usepackage{microtype}

\usepackage{inconsolata}

\usepackage{graphicx}
\usepackage{booktabs}
\usepackage{url}
\usepackage{hyperref}
\hypersetup{
    colorlinks=true,
    linkcolor=blue
}
\usepackage{cleveref}
\usepackage{bbm}

\title{Stronger Re-identification Attacks through Reasoning and Aggregation}

\author{Lucas Georges Gabriel Charpentier \\
  University of Oslo \\
  Language Technology Group \\
  \texttt{lgcharpe@ifi.uio.no} \\\And
  Pierre Lison \\
  Norwegian Computing Center \\ Oslo, Norway \\
  \texttt{plison@nr.no} \\}

\begin{document}

\maketitle
\begin{abstract}
Text de-identification techniques are often used to mask personally identifiable information (PII) from documents. Their ability to conceal the identity of the individuals mentioned in a text is, however, hard to measure. Recent work has shown how the robustness of de-identification methods could be assessed by attempting the reverse process of \textit{re-identification}, based on an automated adversary using its background knowledge to uncover the PIIs that have been masked. This paper presents two complementary strategies to build stronger re-identification attacks. We first show that (1) the \textit{order} in which the PII spans are re-identified matters, and that aggregating predictions across multiple orderings leads to improved results. We also find that (2) reasoning models can boost the re-identification performance, especially when the adversary is assumed to have access to extensive background knowledge.
\end{abstract}

\section{Introduction}

The growing volumes of personal data shared online have brought privacy-enhancing technologies into sharper focus. One important type of privacy-enhancing technology is \textit{text de-identification}, which aims to obscure the occurrence of personally identifiable information (PII) in text documents \cite{Lison2021-bf}. Such PIIs encompass direct identifiers  (such as name, mobile phone numbers or home addresses), but also indirect identifiers (demographic attributes such as age, gender, ethnicity, as well as places, organisations or dates associated with the individual), which are often harder to detect. Important use cases of text de-identification are court judgments \cite{pilan-etal-2022-text,Terzidou2023-gt} and medical records \cite{Neamatullah2008-pu,Kovacevic2024-fn}.

A key challenge in text de-identification lies in how to evaluate the quality of the resulting de-identified texts. One solution, put forward by several authors \cite{Morris2022-oo,manzanares2024evaluating}, is to adopt an adversarial setup that seeks to infer the content of the PIIs that had been concealed. To enhance the effectiveness of such attacks, the adversary can be connected to data sources representing the background knowledge assumed to be available to a motivated intruder \cite{charpentier-lison-2025-identification}. 

\begin{figure}[t]
  \setlength\itemsep{-0.0mm}
    \centering
    \includegraphics[scale=0.23]{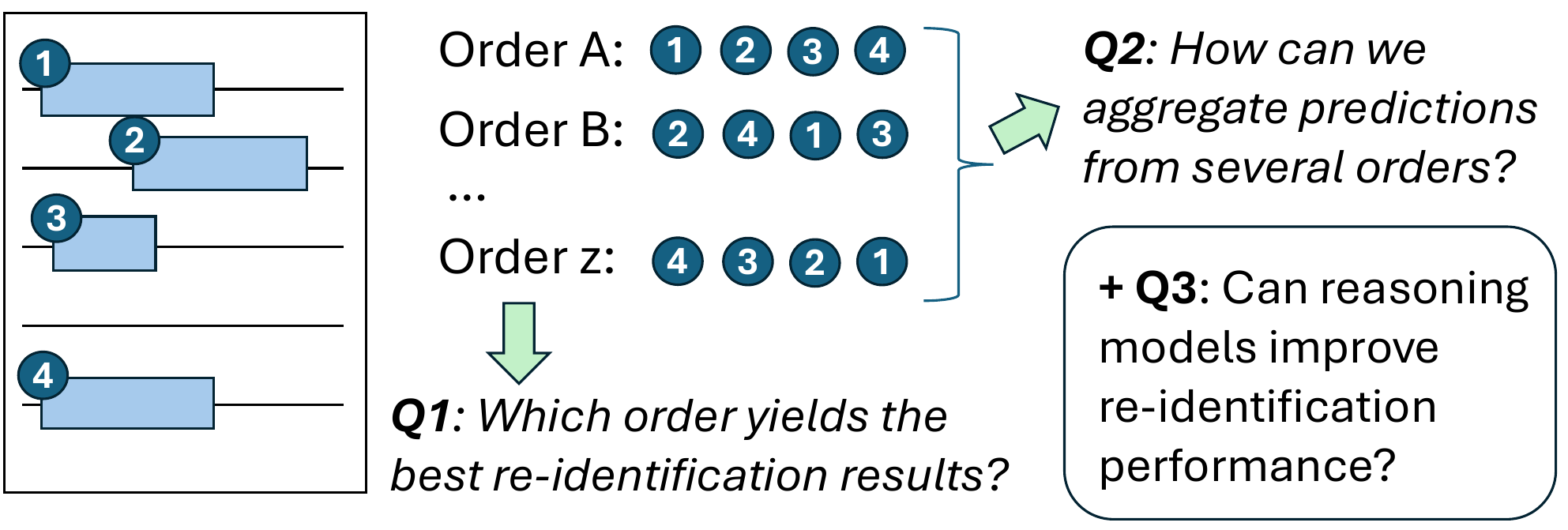}
    \caption{The three questions explored in this paper.} \vspace{-2mm}
    \label{fig:sketch}
\end{figure}

This paper builds upon this adversarial approach and seeks to enhance the strength of the re-identification attacks in two directions:
\begin{enumerate}
    \item We first explore the role played by the \textit{order} in which the masked PIIs are re-identified. We notably find that the aggregation of predictions from multiple orders can substantially improve the re-identification performance.
    \item We also show that LLMs endowed with reasoning abilities can similarly improve the accuracy of the re-identification, although the scale of this improvement depends on the knowledge available to the adversary. 
\end{enumerate}

%The rest of this paper is as follows. 
We provide in Section \ref{sec:background} a short overview on text de-identification and its evaluation through adversarial techniques. Section \ref{sec:methods} details how to order the PIIs to re-identify and aggregate their results. Finally, Section \ref{sec:experiments} evaluates those ordering strategies and the use of reasoning models.

\section{Background}
\label{sec:background}

\paragraph{De-identification} Given a document including personal information, text de-identification aims to detect and mask text spans that reveal Personally Identifiable Information (PII), and thereby conceal the identity of the individuals mentioned in that text. Approaches to this task include rule-based methods \cite{Neamatullah2008-pu}  statistical techniques \cite{Sanchez2016-so}, neural models for sequence labelling \cite{Dernoncourt2017-tk,pilan-etal-2022-text} and LLMs \cite{Liu2023-tm,Larson2024-vx,Staab2024-ju}.

\paragraph{Re-identification} The robustness of those  methods can be tested through the reverse process of \textit{re-identification}, with the aim to infer back the identity of the individuals protected in the document \cite{Mozes2021-xt,manzanares2024evaluating}. \citet{morris-etal-2022-unsupervised} employed Wikipedia infoboxes as  background knowledge to re-identify anonymised Wikipedia articles. This method is subsequently used to enhance de-identification and was recently extended to medical data in \citep{morris2025diri}.  

\citet{charpentier-lison-2025-identification} propose uncovering each masked PII as a preliminary step before attempting to predict the protected individual's identity. Their approach combines a retrieval step -- seeking to find relevant documents in a database of background knowledge -- with an infilling step where an LLM makes use of the retrieved documents to infer possible values for each PII.  The retrieval step is closely inspired by retrieval-augmented models \citep{NEURIPS2020_6b493230,10.5555/3524938.3525306,ram-etal-2023-context} and allows the background knowledge of the adversary to be edited, inspected and updated without requiring any change to the infilling model. The experiments in this paper build upon and further extend this retrieval-based approach.

\section{Methods}
\label{sec:methods}

\subsection{Re-identification model}

We assume an adversary with access to background knowledge deemed relevant for the re-identification task. This background knowledge takes, in practice, the form of a database of text documents related to the domain of the documents to re-identify. 

Based on this background knowledge, the adversary conducts the re-identification in two stages:

\paragraph{Retrieval} Given a de-identified document, a sparse retrieval model is first used to select the top-100 most relevant documents from the background knowledge. The selected documents are then split into chunks. For each masked PII span in the de-identified document, we take a local context window around it (called local context in the rest of this paper). We then pass the chunks and local context to a \textit{dense retriever} to identify the most relevant chunks for the PII span to re-identify. 

Although pre-trained retrieval models could in principle be employed, they perform poorly on this task, as they are often heavily biased towards named entities -- precisely the parts that are masked in de-identified documents. To address this shortcoming, we train a dense retriever using Wikipedia biographies as input and the entire Wikipedia as background knowledge. The positive class corresponds to text chunks that mention the entity to re-identify (or a synonym), while the negative class does not mention that entity. More details on the training of this dense retriever can be found in \Cref{sec:dense_retriever_appendix}.

\paragraph{Infilling} Finally, we give these relevant chunks and local context to an LLM that will seek to \textit{infill} the value of the masked PII span. After re-identifying a span, we substitute its mask with the inferred value in the document and repeat the process until all masked PII spans are restored. 

\subsection{Re-identification order}

A de-identified document often contains numerous masked PIIs. In what sequence should the model attempt to infer their values? Since each re-identification depends on the current state of the document -- including previously re-identified spans -- the order of re-identification can significantly influence the final outcome.

We experimented with 4 alternative orderings:

\paragraph{Top-down} Re-identify the masked PII spans sequentially from the first to the last. \vspace{-1mm}
\paragraph{Bottom-up} Re-identify the PII spans in reverse order, from last to first.  \vspace{-1mm}
\paragraph{Random} Select a PII span at random among the ones not yet re-identified.  \vspace{-1mm}
\paragraph{Entropy-based} Compute the entropy of each PII span and re-identify them in ascending order. The entropy \( H(s) \) of a span \( s \), composed of \( k \) tokens with associated probabilities \( p_1, p_2, \ldots, p_k \) derived from the LLM, is defined as:
\begin{equation}
H(s) = -\sum_{i=1}^{k} p_i \log p_k    
\end{equation}
This ordering corresponds to sorting the PII spans from the ones whose value is easiest to infer (low entropy) to the ones where the model is most uncertain (high entropy). Such easy-to-hard progression mirrors the principles of curriculum learning \cite{Soviany2022-fc}. 

Entropy values are calculated only once at the onset of the re-identification process. While recalculating them at each step is possible (to reflect e.g.~how past decisions may reshuffle the ordering), doing so  also increases computational complexity by $(n+1)/2$, where $n$ is the total number of spans.

\subsection{Aggregation}

Re-identification need not follow one single fixed order. Certain orderings may indeed benefit from positive feedback loops -- where an already re-identified PII span provides useful cues for the remaining ones -- while others may be negatively impacted by the propagation of early errors. 

In light of this observation, we explore aggregation strategies based on \textit{weighted voting}. Specifically, we combine multiple re-identification outputs by summing the probabilities assigned to each span across different runs and selecting the span with the highest total probability. Formally, we first assume a PII span $s$ for which $m$ candidate values $c_1, ..., c_m$ were found by running the re-identification $m$ times with various orderings\footnote{Some of those candidates might be identical.}. Let $p_1, ...p_m$ denote the total probability of those candidates according to the LLM. The aggregated score $A_s(c)$ of a candidate value $c$ for the span $s$ is then computed as:
\begin{equation}
    A_s(c) = \sum_{i=1}^m \mathbbm{1}(c_i = c) \ p_i
\end{equation}
where $\mathbbm{1}$ is the indicator function. The candidate $c$ yielding the highest score $A_s(c)$ is then selected. 

We apply weighted voting over three configurations: (i) 10 randomly sampled orders (\textsc{Random}), (ii) the top-down and bottom-up orders (\textsc{Top-Bottom}), and (iii) all available orders (\textsc{All}).

\section{Experiments}
\label{sec:experiments}

\subsection{Setup}

% As in \citep{charpentier-lison-2025-identification} we use any text chunk that contains either the masked span or a re-direction to the span as a positive retrieved document. We make sure that each span in the training set has at least two positive document (from which one can be the original document). In total we have about 160K local contexts, each with between 2-15 positive and negative retrievals.

\paragraph{Data} We use the test portion of the Text Anonymization Benchmark (TAB) dataset. TAB is a compilation of court cases from the European Court of Human Rights (ECHR) which have been manually de-identified and annotated with various privacy-relevant information \citep{pilan-etal-2022-text}. In particular, each PII span was marked as being either a direct identifier or a so-called \textit{quasi-identifier}\footnote{A quasi-identifier is an information that is not \textit{per se} sufficient to single out the person, but may do so when combined with other quasi-identifiers \cite{Domingo-Ferrer2016-un}}. There are 127 cases in total in the test set. Further details on the data are in \Cref{sec:data_details}.

\paragraph{Background Knowledge} We use two different levels of background knowledge. The first reflects general knowledge available on the web, consisting of court summaries, legal reports and communicated cases. We also use an instruction-tuned version of
Mistral-12B \citep{jiang2023mistral7b} to supplement this collection with a set of synthetic articles and blog posts providing various summaries of the original court cases. The second level of background knowledge is a worst-case scenario mimicking a particularly strong adversary that has access to the original versions of all court judgments.  

\paragraph{Metrics} We assess re-identification performance with two metrics. The first is \textit{exact match accuracy} between the predicted span and the original value. The second is \textit{word-level recall}, measuring the proportion of words in the prediction that also appear in the original span. This accounts for partial matches and simplifications, such as identifying "John Smith" instead of "Mr John Smith".

\begin{table*}[t!]
    \centering
    \begin{tabular}{@{}lccc@{\extracolsep{3em}}c@{\extracolsep{1em}}cc@{}}
        \toprule
        \textbf{Background knowledge: $\rightarrow$} & \multicolumn{3}{c@{\extracolsep{3em}}}{\textbf{General}} & \multicolumn{3}{c@{}}{\textbf{Worst-case}} \\
        \cmidrule(lr){2-4} \cmidrule(lr){5-7}
        \textbf{Ordering strategy} $\downarrow$ & \textbf{Direct} & \textbf{Quasi} & \textbf{All} & \textbf{Direct} & \textbf{Quasi} & \textbf{All} \\
        \midrule
        \small{\textsc{Single}} \\
        \hspace{.5em}Top-down & 6.2 & 10.9 & 10.5 & \underline{75.4} & \underline{42.2} & \underline{44.7} \\
        \hspace{.5em}Bottom-up & \underline{6.8} & 11.8 & 11.4 & 71.7 & 33.9 & 36.8 \\ 
        \hspace{.5em}Random & 5.3 & 12.1 & 11.5 & 67.3 & 34.6 & 37.1 \\
        \hspace{.5em}Entropy-based & 5.7 & \underline{12.7} & \underline{12.1} & 71.8 & 40.3 & 42.8 \\
        \small{\textsc{Weighted Voting}} \\
        \hspace{.5em}Random & 6.1 & \textbf{15.2} & 14.3 & 75.8 & 45.2 & 47.6 \\
        \hspace{.5em}Top-Bottom & 6.6 & 14.4 & 13.7 & \underline{\textbf{77.1}} & 43.4 & 46.1 \\
        \hspace{.5em}All & \underline{\textbf{6.9}} & \underline{\textbf{15.2}} & \underline{\textbf{14.5}} & 76.9 & \underline{\textbf{46.3}} & \underline{\textbf{48.7}} \\
        \bottomrule
    \end{tabular}
    \caption{Exact match performance of the re-identification with the instruction-tuned version of the Qwen3 infilling model. Best scores for each level of knowledge are bolded, and best scores for each sub-class and level of knowledge (single order, weighted voting) are underlined. "Direct" and "Quasi" refer to the distinction between direct and quasi-identifiers as marked by annotators in TAB. All results computed over a single run.}
    \label{tab:instr_res}
\end{table*}

\begin{table*}[t!]
    \centering
    \begin{tabular}{@{}lccc@{\extracolsep{3em}}c@{\extracolsep{1em}}cc@{}}
        \toprule
        \textbf{Background knowledge: $\rightarrow$} & \multicolumn{3}{c@{\extracolsep{3em}}}{\textbf{General}} & \multicolumn{3}{c@{}}{\textbf{Worst-case}} \\
        \cmidrule(lr){2-4} \cmidrule(lr){5-7}
        \textbf{Ordering strategy} $\downarrow$& \textbf{Direct} & \textbf{Quasi} & \textbf{All} & \textbf{Direct} & \textbf{Quasi} & \textbf{All} \\
        \midrule
        \small{\textsc{Single}} \\
        \hspace{.5em}Top-down & 6.1 & 13.7 & 12.9 & \underline{81.3} & 49.0 & 51.3 \\
        \hspace{.5em}Bottom-up & 5.1 & 14.0 & 13.2 & 76.1 & 52.3 & 53.9 \\ 
        \hspace{.5em}Random & \underline{\textbf{7.5}} & 14.7 & 14.1 & 76.6 & 51.7 & 53.7 \\
        \hspace{.5em}Entropy-based & 5.9 & \underline{15.4} & \underline{14.5} & 79.8 & \underline{52.8} & \underline{55.0} \\
        \small{\textsc{Weighted Voting}} \\
        \hspace{.5em}Top-Bottom & 6.3 & 14.6 & 13.9 & 78.6 & 52.0 & 53.9 \\
        \hspace{.5em}All & \underline{6.6} & \underline{\textbf{16.2}} & \underline{\textbf{15.3}} & \underline{\textbf{81.5}} & \underline{\textbf{55.2}} & \underline{\textbf{57.2}} \\
        \bottomrule
    \end{tabular}
    \caption{Exact match performance of the re-identification with the reasoning-optimised version of the Qwen3 infilling model. Best scores for each level of knowledge are bolded, and best scores for each sub-class and level of knowledge (single order, weighted voting) are underlined. All results computed over a single run}
    \label{tab:reason_res}
\end{table*}

\subsection{Implementation}

\paragraph{Retrieval} The sparse retrieval is implemented with a BMx model \citep{li2024bmxentropyweightedsimilaritysemanticenhanced}, a variation of the BM25 model \citep{robertson2009probabilistic} which accounts for both lexical and semantic similarities. The dense retrieval relies on a  ColBERT-style retriever \citep{khattab2020colbert} initialised with ModernBERT-base \citep{warner-etal-2025-smarter} for both the document and query encoder.

\paragraph{Infilling} We use Qwen3 4B \citep{qwen3technicalreport} in two distinct configurations: \textit{\href{https://huggingface.co/Qwen/Qwen3-4B-Instruct-2507}{instruction-tuned}} or \textit{\href{https://huggingface.co/Qwen/Qwen3-4B-Thinking-2507}{reasoning-optimised}}. To re-identify a PII span, we take its local context along with the 10 most relevant chunks provided by the retrieval module, and pass them to the infilling model. The prompt first includes the 10 retrieved chunks, followed by simple instructions to give the value of the token marked with a \verb|[MASK]| token. All other de-identified spans are marked with \verb|anon|. The prompt and additional details are provided in \Cref{subsec:infilling}.

% \begin{description}
%     \setlength{\itemsep}{0pt}
%     \item[Top-down] We re-identify the spans starting with the first and finishing with the last span found in the text.
%     \item[Bottom-up] We re-identify the spans starting from the last one and finishing with the first found in the text.
%     \item[Random] We randomly choose which span to re-identify at each pass of the re-identification model.
%     \item[Entropy-based] We first calculate the sum log-likelihood of the re-identification for each span. We then order them from the highest sum log-likelihood to the lowest and re-identify the text in that order.
%     \item[Weighted Voting] We aggregate multiple re-identifications by summing together the sum probabilities of each span from each re-identification and then choosing the span with the highest sum probability as the re-identification. We do this on either 10 random orders (Random), the top-down and bottom-up orders (Top-Bottom), or all orders (All).
% \end{description}

\subsection{Results}
\label{sec:results}

The re-identification results with the instruction-tuned model are given in \Cref{tab:instr_res}. Overall, the choice of re-identification order seems to have a negligible impact on the exact match accuracy, although the entropy-based strategy performs relatively well across the board.  We also observe that aggregating results over several orders does improve the re-identification performance, particularly for quasi-identifiers. Similar trends are found in the word recall performance (\Cref{tab:instr_wr_res}). 

%The general scenario results in  shows that the differences between each order in the general background setting are very minimal. We see some small improvement in the overall accuracy when using the Entropy-based ordering, but this is probably due to random variation. In the worst-case scenario , we see some differences emerge. We see that the Top-down order performs the best both with direct and quasi-identifiers. This could indicate that, depending on the level of knowledge, the importance of ordering changes. We see very similar trends for the word recall performance (\Cref{tab:instr_tr,tab:instr_tr_l4} found in \Cref{sec:word_recall}) except for direct-identifiers in the worst case scenario, where the Bottom-up order slightly edges out the Top-down order.

%Focusing on the aggregation results for the general #scenario from \Cref{tab:instr_res}, we see that aggregating 10 random orders significantly improves overall performance (about a three percentage point increase). However, the improved performance comes mostly from better identification of quasi-identifiers rather than overall increases. This could mean that direct identifiers tend to be re-identified similarly, no matter the order. We see similar trends when aggregating the Top-down and Bottom-up orders. Finally, by aggregating all orders, we slightly outperform all other aggregations (about one percentage point). Similar trends can be seen in the worst case scenario and for the word recall performance (\Cref{tab:instr_tr,tab:instr_tr_l4}). 

When applying the reasoning version of the infilling model (\Cref{tab:reason_res}), we observe a substantial boost in the overall re-identification performance compared to the results obtained with the instruction-tuned model\footnote{Note, however, that the reasoning model is more computationally demanding, with inference times on average 25 times greater than for its instruction-tuned counterpart.}. The entropy-based order also seems to perform well in this setting. This increase is particularly prominent when considering an adversary with access to extensive background knowledge. The results obtained for word recall show similar trends (\Cref{tab:reason_wr_res} found in \Cref{sec:word_recall}). 

%When considering the worst-case background, we see a larger increase in performance from reasoning both on single and aggregated orders (between 5 and 10 percentage point increase overall). In this case, we do observe an increase in the direct-identifier re-identification; however, most of the improvement comes from the quasi-identifier re-identification. We see similar trends for the word recall performance (\Cref{tab:reason_tr_l4} found in \Cref{sec:word_recall}).

\section{Conclusion}

Re-identification attacks are increasingly used to test the robustness of text de-identification methods. This paper proposed a set of practical techniques for building stronger re-identification attacks. In particular, we show that reasoning-optimised LLMs are able to enhance re-identification performance, particularly when coupled to adversaries with access to rich background knowledge. The paper also explored the role of the re-identification order. Although entropy-based orderings only yielded modest performance gains, the aggregation of predictions over multiple orders was shown to enhance the re-identification accuracy.

%Overall, we observe four different trends from our results. The first is that using reasoning models tends to improve performance for re-identification when using a single re-identification. Further, we notice that doing an aggregation of results leads to significantly better results than using a single order. Next, we see that the more relevant the background, the larger the gains from reasoning or aggregation become. Finally, we see that most of the improvements come from better quasi-identifier re-identification, rather than overall re-identification. The first three trends are expected, given both the complexity of the task and the importance of background information. The final trend is more surprising. We posit that the reason behind it is that the context around quasi-identifiers can lead to multiple possible re-identifications, making it harder to re-identify.

%Through this paper, we have shown that re-identification order importance depends on the level of background knowledge. As the background becomes more relevant, so does the importance of the re-identification order. We also show that aggregating multiple orders together can lead to stronger re-identification. Finally, we show that using reasoning models improves performance; however, the extent of the improvement depends on the relevance of the background. This suggests that reasoning models can make more robust adversaries to sanitisation and help alleviate the risk of re-identification when used to improve sanitisation.

\section*{Limitations}

Due to computing constraints, we only tested a single model architecture and size for the re-identification experiments. While alternative architectures or model sizes might lead to different outcomes, we believe most setups would benefit from the use of reasoning models and aggregating predictions over different runs -- although the performance gains might diminish with model size. We also only evaluate the re-identification in a zero-shot setting, due to the difficulty of providing relevant few-shot examples in this retrieval-based scheme. 

We only consider English texts as background knowledge for the re-identification. It is possible that using background knowledge expressed in different languages could boost re-identification. This would, however, require adaptation of the sparse retrieval since the lexical performance would be near zero for texts in different languages. We also believe that integrating alternative data types -- such as tables or knowledge graphs -- could benefit the task and  affect the impact of re-identification order.

\section*{Ethical Statement}

Although our re-identification work could potentially be exploited to extract personal information, its primary aim is to develop robust adversarial models that enhance sanitisation -- similar to red teaming practices in cybersecurity. By pairing re-identification models with sanitisation systems, we can strengthen the latter’s resilience and assess the extent of background knowledge they can withstand.

The two datasets used in this paper (TAB and Wikipedia) are publicly available (articles on Wikipedia are derived from public information on people, and the TAB dataset is created from court cases from the ECHR).

\bibliography{latex/custom,latex/anthology11,latex/anthology12}

\newpage 

\appendix

\section{Technical details}
\label{sec:details}

\Cref{tab:models} details all the models used in this paper, as well as what they were used for, their size, license, a link to the model, and the citation. All models were used in accordance with their respective licenses and intended uses. All models except the Spacy NER model are downloaded from HuggingFace \citep{wolf-etal-2020-transformers}.

\subsection{Dense retriever}
\label{sec:dense_retriever_appendix}

To use the Wikipedia dataset for training our dense retriever, we start by de-identifying our biographies section of the dataset using an NER model to detect entities to mask. We then set the non-biography portion to be the background knowledge. Once the sparse retriever has selected the top documents, these are split into chunks of 600 characters. They are then labelled as positive if the chunk contains the masked entity (or a synonym), and negative otherwise. For each class, up to 15 text chunks are selected (with a minimum of two positive examples), resulting in a dataset comprising approximately 160\,000 local texts paired with their corresponding positive and negative chunks.

\Cref{tab:retr-details} shows the training details for the dense retriever model. The training took around 3 GPU-hours to complete using an RTX 3090 with PyTorch 2.8 and Python 3.13.  During training, all spans are de-identified.

\begin{table}[ht]
    \centering
    \begin{tabular}{@{}lc@{}}
        \toprule
        \textbf{Hyperparameter} & \textbf{Value} \\
        \midrule
        Learning Rate & $10^{-5}$ \\
        Scheduler & Cosine decay \\
        Batch Size & 128 \\
        \# Steps & 10\,000 \\
        \# Warmup Steps & 1\,000 \\
        Compression Size & 64 \\
        Seed & 42 \\
        Query Encoder & \texttt{ModernBERT-base} \\
        Document Encoder & \texttt{ModernBERT-base} \\
        \bottomrule
    \end{tabular}
    \caption{Training details for the ColBERT models}
    \label{tab:retr-details}
\end{table}

As in the ColBERT paper, we compress the final logits to a hidden size of 64. \citet{khattab2020colbert} showed that doing this compression leads to small performance losses but large memory gains. For the retrieved chunks, we chunk documents into lengths of approximately 1\,200 characters (600 during training). For the query, we use a context window of 1\,000 characters (200 during training). For the model to know which span the retrieval is on, we mark it with the \verb|[MASK]| token, and mark the other masked spans with a \verb|[ANON]| token.

\subsection{Infilling}
\label{subsec:infilling}

For the single random order, we used seed 42 for both levels of background knowledge and re-identifier models. For the aggregation of random orders with the instruction model, we used seeds 0 to 10 for the random orders.

The Top-down, Bottom-up, and Random orders each took about 2h to fully re-identify the TAB dataset when using the instruction-tuned model. The Entropy-based order took 3 hours. For the reasoning model, the Top-down, Bottom-up, and Random orders took 50 hours each, while the Entropy-based order took 75 hours. In total, the inference took 508 GPU-hours using a MI250X GPU.
We used PyTorch 2.7 and VLLM 0.9.1 for inference.

\begin{table*}[ht]
    \small
    \centering
    \begin{tabular}{@{}llllll@{}}
    \toprule
        \textbf{Model} & \textbf{Use} & \textbf{\# Params} & \textbf{License} & \textbf{Link} & \textbf{Citation} \\
        \midrule
         \texttt{en\_core\_web\_trf} (RoBERTa-base) & NER model & 125M & MIT & \href{https://spacy.io/}{Link} & \citep{Honnibal_spaCy_Industrial-strength_Natural_2020} \\
         \texttt{Mistral-Nemo-Instruct-2407} & Generating articles & 12.2B & Apache 2 & \href{https://huggingface.co/mistralai/Mistral-Nemo-Instruct-2407}{Link} & \citep{jiang2023mistral7b} \\
         \texttt{ModernBERT-base} & Dense Retriever & 150M & Apache 2 & \href{https://huggingface.co/answerdotai/ModernBERT-base}{Link} & \citep{warner-etal-2025-smarter} \\
         \texttt{Qwen3-4B-Instruct-2507} & Re-identifier & 4B & Apache 2 & \href{https://huggingface.co/Qwen/Qwen3-4B-Instruct-2507}{Link} & \citep{qwen3technicalreport} \\
         \texttt{Qwen3-4B-Thinking-2507} & Re-identifier & 4B & Apache 2 & \href{https://huggingface.co/Qwen/Qwen3-4B-Thinking-2507}{Link} & \citep{qwen3technicalreport} \\
         \bottomrule
    \end{tabular}
    \caption{Details on models used.}
    \label{tab:models}
\end{table*}

The prompt we used for re-identification is the following:

\begin{verbatim}
Given the following potentially 
relevant passages:

{retrieved}

Re-identify the fill in the blank
(marked with [MASK]) in the text
below. Structure the answer as a
JSON file with the field
're-identification' containing the
value of [MASK]:

{text}

Answer:
\end{verbatim}

The \verb|{retrieved}| are the top-10 retrieved chunks by our dense retriever, separated by two new lines. The \verb|{text}| is the masked text span and its local context.

\section{Dataset details}
\label{sec:data_details}

\subsection{Wikipedia}

We only use the English version of Wikipedia, which uses the CC BY-SA license. We do not make any edits to the data outside of splitting it into biographies and non-biographies. The dataset contains PIIs of famous/known people and entities. Given that our work involves measuring re-identification and the data is openly available to everyone, we do not anonymise the PIIs.

% Although pretrained retrieval models could in principle be employed for this task, they perform poorly on this task, as they are often heavily biased towards named entities -- precisely the parts that are masked in de-identified documents. To this end, we train a dense retriever using Wikipedia biographies as input and the entire Wikipedia as the background knowledge. De-identification of the biographies is performed using an NER model to detect entities to mask. Documents selected by the sparse retriever are segmented into chunks, and each chunk is then labelled as positive if it contains the value (or a synonym) of the PII span, and negative otherwise. We ensure that each local context includes at least two positive documents. For each class, up to 15 text chunks are selected, resulting in a dataset comprising approximately 160\,000 local texts paired with their corresponding positive and negative chunks.

\subsection{TAB}

For the TAB dataset, we only use the test split of the dataset.  As there may be multiple annotation layers for a given case, we select the de-identification of a random annotator to be our de-identified case. 
This data is shared under the MIT license. This data represents court cases brought to the European Court of Human Rights (ECHR); as such, they contain various PIIs. As the data is publicly available and our usage requires these PIIs, we do not anonymise them. 

For the background knowledge, we use communicated cases, reports, and legal summaries found on the \href{https://hudoc.echr.coe.int/}{HUDOC} database of the ECHR. We make sure that all the retrieved data is in English. As some of the court reports from the ECHR are written in French, we translate those to English using the \href{https://huggingface.co/Unbabel/Tower-Plus-2B}{Tower-Plus-2B} model from Unbabel \cite{rei2025towerplus}. 

\section{Word Recall Results}
\label{sec:word_recall}

\Cref{tab:instr_wr_res} shows the word recall performance of the instruction-tuned version of the Qwen3 infilling model. \Cref{tab:reason_wr_res} shows the word recall performance of the reasoning-optimised version of the Qwen3 infilling model.

\begin{table*}[t!]
    \centering
    \begin{tabular}{@{}lccc@{\extracolsep{3em}}c@{\extracolsep{1em}}cc@{}}
        \toprule
        \textbf{Background knowledge: $\rightarrow$} & \multicolumn{3}{c@{\extracolsep{3em}}}{\textbf{General}} & \multicolumn{3}{c@{}}{\textbf{Worst-case}} \\
        \cmidrule(lr){2-4} \cmidrule(lr){5-7}
        \textbf{Ordering strategy} $\downarrow$ & \textbf{Direct} & \textbf{Quasi} & \textbf{All} & \textbf{Direct} & \textbf{Quasi} & \textbf{All} \\
        \midrule
        \small{\textsc{Single}} \\
        \hspace{.5em}Top-down & 13.3 & 19.0 & 18.2 & 79.7 & \underline{49.3} & \underline{50.7} \\
        \hspace{.5em}Bottom-up & \underline{14.5} & 20.5 & 19.9 & \underline{80.4} & 43.0 & 45.3 \\ 
        \hspace{.5em}Random & 12.1 & 20.8 & 19.8 & 75.8 & 44.1 & 45.9 \\
        \hspace{.5em}Entropy-based & 13.2 & \underline{21.3} & \underline{20.5} & 78.8 & 48.1 & 49.6 \\
        \small{\textsc{Weighted Voting}} \\
        \hspace{.5em}Random & 13.9 & 24.8 & 23.8 & 82.1 & 55.0 & 56.8 \\
        \hspace{.5em}Top-Bottom & 14.6 & 23.6 & 22.7 & 82.6 & 53.1 & 54.8 \\
        \hspace{.5em}All & \underline{\textbf{15.2}} & \underline{\textbf{25.1}} & \underline{\textbf{24.2}} & \underline{\textbf{82.7}} & \underline{\textbf{56.3}} & \underline{\textbf{58.1}} \\
        \bottomrule
    \end{tabular}
    \caption{Word recall performance of the re-identification with the instruction-tuned version of the Qwen3 infilling model. Best scores for each level of knowledge are bolded, and best scores for each sub-class and level of knowledge (single order, weighted voting) are underlined. "Direct" and "Quasi" refer to the distinction between direct and quasi-identifiers as marked by annotators in TAB. All results computed over a single run.}
    \label{tab:instr_wr_res}
\end{table*}

\begin{table*}[t!]
    \centering
    \begin{tabular}{@{}lccc@{\extracolsep{3em}}c@{\extracolsep{1em}}cc@{}}
        \toprule
        \textbf{Background knowledge: $\rightarrow$} & \multicolumn{3}{c@{\extracolsep{3em}}}{\textbf{General}} & \multicolumn{3}{c@{}}{\textbf{Worst-case}} \\
        \cmidrule(lr){2-4} \cmidrule(lr){5-7}
        \textbf{Ordering strategy} $\downarrow$ & \textbf{Direct} & \textbf{Quasi} & \textbf{All} & \textbf{Direct} & \textbf{Quasi} & \textbf{All} \\
        \midrule
        \small{\textsc{Single}} \\
        \hspace{.5em}Top-down & 13.6 & 24.4 & 23.7 & \underline{86.5} & 60.6 & 62.2 \\
        \hspace{.5em}Bottom-up & 9.7 & 24.7 & 23.7 & 84.3 & 64.3 & 65.4 \\ 
        \hspace{.5em}Random & \underline{\textbf{16.0}} & 24.9 & 24.2 & 82.7 & 64.0 & 65.1 \\
        \hspace{.5em}Entropy-based & 14.1 & \underline{25.8} & \underline{25.0} & 86.4 & \underline{64.6} & \underline{66.0} \\
        \small{\textsc{Weighted Voting}} \\
        \hspace{.5em}Top-Bottom & \underline{14.8} & 27.4 & 26.5 & \underline{\textbf{87.8}} & 67.1 & 68.4 \\
        \hspace{.5em}All & 14.7 & \underline{\textbf{27.7}} & \underline{\textbf{26.9}} & 87.1 & \underline{\textbf{67.6}} & \underline{\textbf{68.9}} \\
        \bottomrule
    \end{tabular}
    \caption{Word performance of the re-identification with the reasoning-optimised version of the Qwen3 infilling model. Best scores for each level of knowledge are bolded, and best scores for each sub-class and level of knowledge (single order, weighted voting) are underlined. "Direct" and "Quasi" refer to the distinction between direct and quasi-identifiers as marked by annotators in TAB. All results computed over a single run.}
    \label{tab:reason_wr_res}
\end{table*}

\end{document}